\documentclass[letterpaper,10pt,journal,twoside]{IEEEtran}

\usepackage{amsmath,amsfonts}
\usepackage{algorithmic}
\usepackage{algorithm}
\usepackage{array}
\usepackage[caption=false,font=normalsize,labelfont=sf,textfont=sf]{subfig}
\usepackage{textcomp}
\usepackage{stfloats}
\usepackage{url}
\usepackage{verbatim}
\usepackage{graphicx}
\usepackage{cite}
\usepackage{xcolor}
\usepackage{svg}

\usepackage{booktabs}
\usepackage{longtable}
\usepackage{tabularx}
\usepackage{caption}
\renewcommand{\tablename}{TABLE}

\usepackage{capt-of}

\begin{document}

\title{GTA-2: A Multi-VLM Framework for Synthesizing   Robot Manipulation Skills via Grounded Task Axes}

\author{
M. Yunus Seker\textsuperscript{1},
Shobhit Aggarwal\textsuperscript{1},
Ruwan Wickramarachchi\textsuperscript{2},
Jonathan Francis\textsuperscript{2},
and Oliver Kroemer\textsuperscript{1}%
\thanks{\textsuperscript{1}The Robotics Institute, Carnegie Mellon University, Pittsburgh, PA, USA
\{mseker; shobhita; okroemer\}@andrew.cmu.edu.}%
\thanks{\textsuperscript{2}Bosch Research, Pittsburgh, PA, USA
\{ruwan.wickramarachchi; jon.francis\}@us.bosch.com.}%
}



\IEEEaftertitletext{%
    \vspace{-1.5\baselineskip}
    \begin{minipage}{\textwidth}
        \centering
        \includegraphics[width=\linewidth]{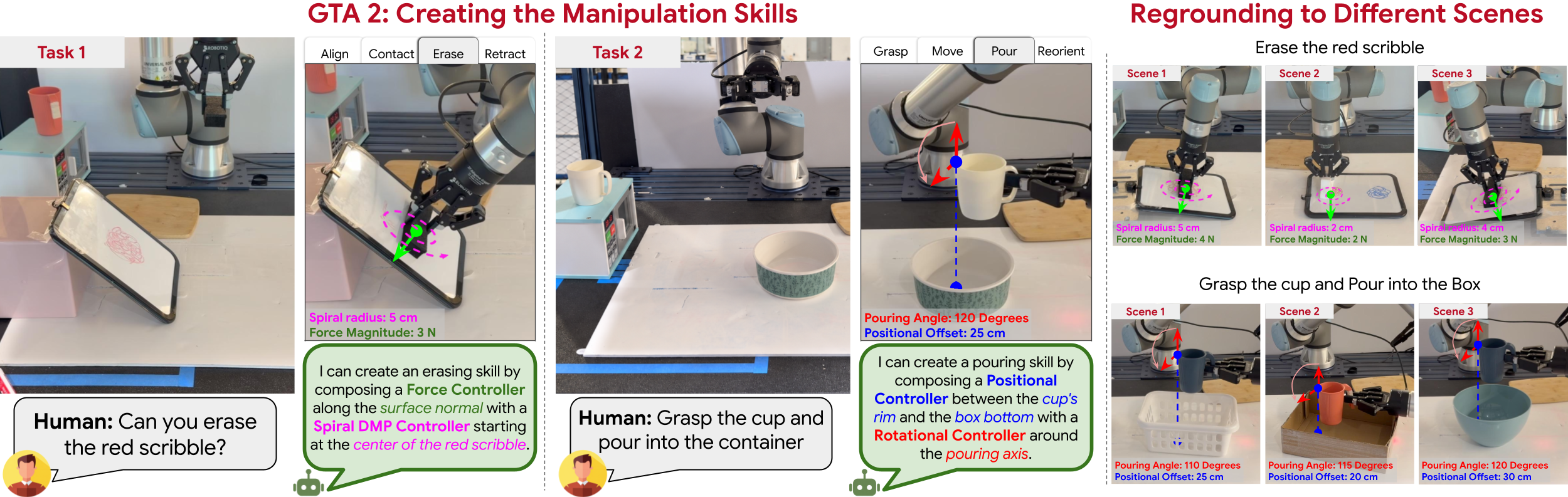}
        \captionof{figure}{\textbf{GTA-2 turns task instructions into reusable, scene-grounded manipulation skills.} (Left) GTA-2 identifies task-relevant object keypoints and axes, and composes parameterized controllers to construct executable skills, illustrated for erasing and pouring. (Right) Across new scene configurations, the controller composition remains fixed while object features and execution parameters are re-grounded for the current observation.}
        \label{fig:teaser}
    \end{minipage}
}

\maketitle

\begin{abstract}
Robotic manipulation tasks are often decomposed into behaviors or skills. However, one often needs to predefine these behaviors for specific tasks or try to  cover a wide range of tasks using generic skills. As a result, these behaviors can remain too coarse to expose the geometric, control, and scene-dependent decisions required for execution. We introduce \textit{Grounded Task Axes v2} (GTA-2), a modular multi-VLM framework that constructs executable, task-bespoke manipulation skills from reusable object-centric task-axis components. Rather than predicting actions end-to-end or composing fixed task-level primitives, GTA-2 represents each skill as semantic subtasks comprising task-relevant keypoints and axes, controller compositions, and scene-dependent parameters. Four specialized VLM agents separately decompose the task, construct an abstract task-axis skill, assign controller parameters, and ground the required visual features from RGB-D observations. This abstraction-to-grounding factorization enables zero-shot skill generation without task-specific robot demonstrations, policy training, or fine-tuning. It also keeps intermediate decisions explicit, allowing targeted human feedback to refine an incorrect stage while preserving correct components. We evaluate GTA-2 on 14 real-robot manipulation tasks against a VLA policy $\pi_{0.5}$ and two Code-as-Policies baselines using task-axis controllers or conventional robot primitives. GTA-2 achieves an average zero-shot success rate of $73.9\%$, exceeding the strongest baseline by $31.4$ percentage points, while targeted refinement raises GTA-2's average success rate to $90.7\%$. \textbf{Project page:}
\url{https://gta2-project.github.io/}
\end{abstract}

\vspace{-4mm}
\section{Introduction}
Robotic manipulation tasks are commonly represented as compositions of simpler behaviors or skills, allowing complex tasks to be decomposed into more manageable units such as reaching, grasping, moving, inserting, or placing. However, these generic skill-level units often remain relatively coarse: each behavior must still encode how the robot should interact with the objects in the current scene, including which object parts are relevant, how the end-effector should be oriented, and what geometric or contact constraints should be satisfied. Natural-language instructions specify task-level intent, but do not by themselves determine these scene-specific geometric and control choices. Bridging this gap requires an intermediate representation that is abstract enough to transfer across object instances and configurations, yet explicit enough for low-level robot controllers to instantiate and execute.

We argue that such a representation requires decomposing behaviors one level further, into reusable object-centric geometric features and controller-level components that can be recombined as needed. Figure~\ref{fig:teaser} illustrates this finer-grained representation through erasing and pouring. For erasing, a spiral-trajectory controller is combined with force control and grounded relative to the scribble center and board normal; for pouring, positional and rotational controllers coordinate the cup with the target container. Once constructed, these controller compositions can be reused across scene configurations, while keypoints, axes, and execution parameters are re-grounded for the current observation. This separation preserves the reusable structure of a manipulation skill while adapting its geometric and control details to each scene.

Recent language-conditioned manipulation systems offer two prominent routes for connecting task intent to robot execution. Vision-language-action (VLA) models learn generalist behaviors that map visual observations and natural-language commands directly to robot actions \cite{shridhar2023perceiver,brohan2022rt,zitkovich2023rt,team2024octo,intelligence2025pi_}. While these systems have demonstrated broad task capabilities, the internal structure of the learned behavior remains largely implicit, making it difficult to identify how the system decomposes a task or reasons about the scene and execution. Language-model-based planning approaches such as Code-as-Policies instead generate executable robot programs by composing available APIs and predefined skill primitives \cite{liang2023code}. However, these primitives typically encapsulate substantial perception, grounding, and control logic, requiring the available skill library to already contain behaviors capable of solving the task. Thus, despite producing different outputs, both approaches largely treat behaviors as atomic units whose internal geometric and control structure is either implicit or predefined. In contrast, we aim to operate at a \textbf{sub-atomic level}, constructing new behaviors online from reusable object-centric features and controller-level components.

Task-axis controllers provide a natural representation for operating at this finer granularity by expressing manipulation behaviors as structured compositions of reusable controllers grounded in object-centric coordinate frames \cite{ballard1984task,mason1981compliance,raibert1981hybrid}. These controllers express motion, orientation, force, and constraint objectives relative to semantically meaningful keypoints and axes, such as a handle axis, a surface normal, an insertion direction, or a grasp point~\cite{manschitz2020learning, migimatsu2020object, king2016rearrangement, berenson2011task,sharma2020learning,sharma2021generalizing}. Grounded Task-Axes (GTA) \cite{seker2025grounded} builds on this idea by representing manipulation skills through task-relevant keypoints, axes, controller compositions, priorities, and action spaces. This representation makes the structure of the skill explicit: an abstract skill can specify what must be achieved in terms of object-relative geometry, and the same skill can then be instantiated in a new scene by grounding the required keypoints and axes. However, prior task-axis-based systems relied on hand-engineered task decompositions, controller selections, and task-axis tuning.

In this work, we introduce \textit{Grounded Task Axes version 2} (GTA-2), a modular multi-VLM framework that automates the construction and grounding of task-axis skills from a high-level task prompt and a visual observation of the scene. Using the semantic and visual reasoning capabilities of VLMs, GTA-2 decomposes the task, constructs abstract controller recipes, assigns execution parameters, and grounds the required object keypoints and axes in the observed scene. Rather than generating actions end-to-end, GTA-2 generates an explicit task-axis skill representation that connects language-level task reasoning to controller-level robot execution.

GTA-2 realizes the separation between abstract skill synthesis and scene-specific grounding through a pipeline of four specialized VLM agents. The first two agents construct the abstract skill representation. A \textbf{Task Decomposer} converts the user instruction into an ordered sequence of subtasks, and a \textbf{Skill Generator} maps each subtask to an abstract skill recipe by selecting the required controller composition, keypoints, task axes, and parameter placeholders. The remaining two agents instantiate this abstract representation in the observed scene. A \textbf{Parameter Setter} assigns task- and scene-specific controller parameters such as offsets, forces, and orientation targets, while a \textbf{Vision Module} grounds the requested keypoints and axes in the RGB-D observation. The resulting grounded skill is then compiled into an executable robot script.

This separation also makes the robot's intermediate decisions explicit, modular, and individually refinable.  Because GTA-2 exposes the subtask sequence, skill recipe, controller parameters, and visual groundings as separate intermediate outputs, a human can provide targeted feedback to any stage of the pipeline. For example, if the task should be decomposed differently, the user can refine the Task Decomposer output; if the robot should use a different contact point or motion constraint, the user can provide feedback to the Skill Generator or Vision Module; and if the robot should lift an object higher before moving it, the correction can be sent to the Parameter Setter. This allows the system to preserve correct parts of the generated skill while updating only the components that require modification, providing a practical interface for human-guided adaptation without retraining an end-to-end policy.

\begin{figure*}
    \centering
    \includegraphics[width=\linewidth]{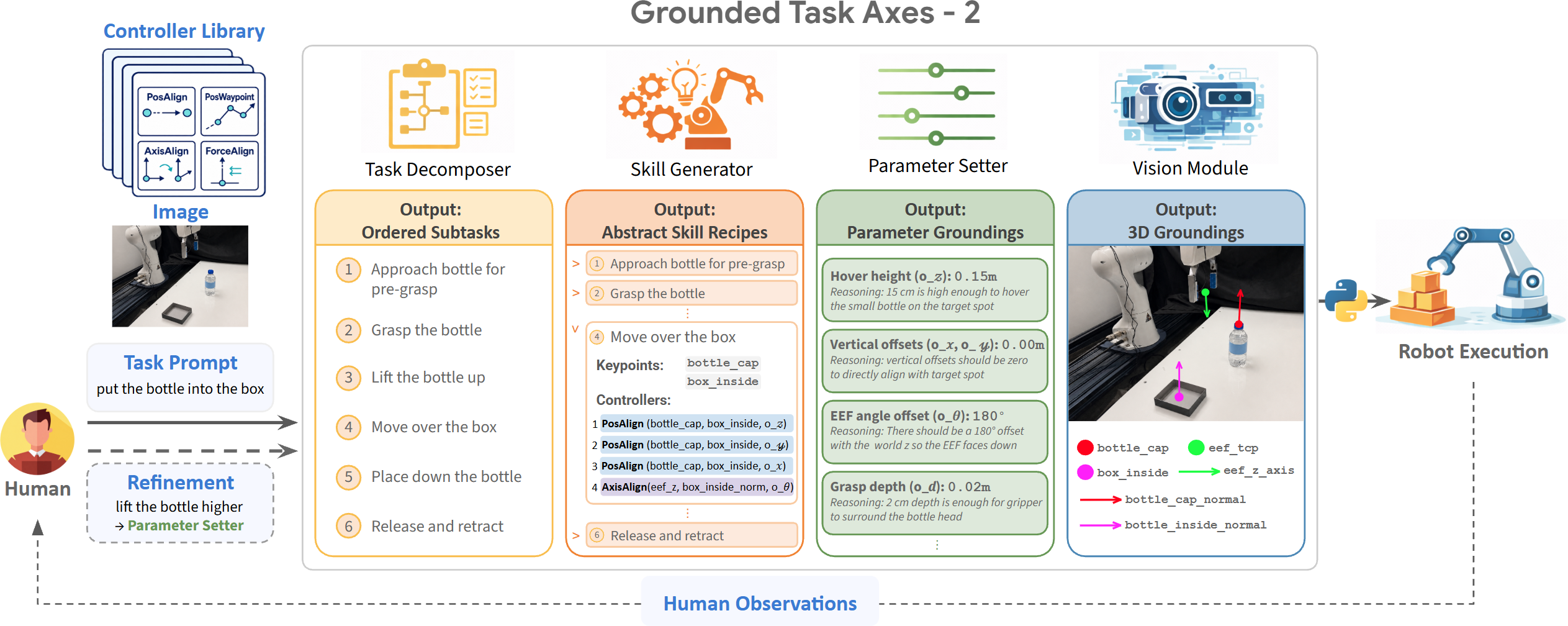}
    \caption{\textbf{Overview of GTA-2.} GTA-2 takes a task prompt, scene image, and controller library as inputs. Specialized agents process these inputs to decompose the task, generate abstract skill recipes, set task-specific parameters, and produce 3D visual groundings. These outputs are combined into an executable robot script. Human feedback provides a refinement loop for updating the subtasks, parameters, or groundings when needed.}
    \vspace{-4mm}
    \label{fig:method_overview}
\end{figure*}

We evaluate GTA-2 on a real robot across 14 diverse manipulation tasks and compare it against three baselines: the generalist VLA policy $\pi_{0.5}$, a Code-as-Policies baseline with access to the same task-axis controllers as GTA-2, and a Code-as-Policies baseline using conventional robot motion and force primitives. Our experiments evaluate zero-shot task execution, generalization of generated skills across scene configurations, and targeted human refinement, while the baseline comparisons isolate the benefits of structured skill generation and task-axis controller composition.

\vspace{-2mm}
\section{Related Work}
\paragraph*{\textbf{Task-Axis Controllers}} 
Task-axis controllers build on a long line of work that represents manipulation in task-relative coordinate systems rather than as generic joint-space motions or single global world frames. Early task-frame and compliance-control formulations showed that manipulation objectives can be specified relative to task-relevant frames and that different task directions can be regulated through position, force, or compliance objectives \cite{ballard1984task,mason1981compliance,raibert1981hybrid}. Subsequent manipulation-planning work used related ideas to define constrained task-space regions, object-centric action spaces, and object-relative planning frames for pose-constrained manipulation, rearrangement, and dynamic environments \cite{berenson2011task,king2016rearrangement,migimatsu2020object}. More recent work has explored learning and composing modular object-centric controllers and sequential force-interaction skills, demonstrating that structured controller representations can improve generalization across objects, subtasks, and interaction phases \cite{manschitz2020learning,sharma2020learning}. Task-axis controllers make this structure explicit by parameterizing motion, orientation, force, and constraint objectives with object keypoints and axes; \cite{sharma2021generalizing} showed that keypoint-grounded task-axis controllers can generalize across variations in object shape, size, and geometry. Grounded Task-Axes (GTA-1) further extended this formulation to zero-shot semantic skill transfer by using visual foundation models to identify corresponding grounding features on novel objects \cite{seker2025grounded}. Building on this line of work, GTA-2 addresses the dependence on human-designed task decompositions, controller recipes, task-axis definitions, parameter choices, and example skills by using a VLM pipeline to generate these directly from a task prompt and scene.

\paragraph*{\textbf{LLM-based plan and code generation}}
LLMs have also been used for robot planning and program generation by translating natural-language goals into sequences of high-level plans or executable code. Systems such as SayCan use LLMs for semantic task reasoning while grounding decisions through affordance models or scene feedback \cite{ichter2023saycan,huang2023inner}. Related approaches, including ProgPrompt and Code-as-Policies \cite{singh2023progprompt,liang2023code}, constrain LLM outputs through program-like prompts, function libraries, perception APIs, or robot-control primitives to produce executable plans or policy code \cite{vemprala2024chatgpt,mu2024robocodex,burns2024genchip,xie2025robopro}. These works show that LLMs can provide useful semantic reasoning for robot behavior generation, but they generally depend on a single-step-generated program using predefined primitives or perception APIs to bridge language and execution. Our work builds on the same generative insight while shifting the focus from directly coding over existing primitives to generating an explicit task-axis skill representation that can be grounded in the scene and refined through feedback.

\paragraph*{\textbf{Vision Language Action models}}
Vision-language-action (VLA) models have recently become a major direction for language-conditioned robotic manipulation, mapping visual observations and natural-language instructions directly to robot actions. Systems such as Perceiver-Actor\cite{shridhar2023perceiver}, RT-1\cite{brohan2022rt}, RT-2\cite{zitkovich2023rt}, Octo\cite{team2024octo}, OpenVLA\cite{kim2025openvla}, $\pi_0$\cite{black2025pi0}, and $\pi_{0.5}$\cite{intelligence2025pi_} show that scaling robot policy learning with transformer architectures, large robot datasets, vision-language pretraining, and cross-embodiment data can produce increasingly capable generalist robot policies. Some recent models, such as $\pi_{0.5}$, also begin to introduce higher-level reasoning by predicting intermediate subgoals before low-level action execution. Nevertheless, the core formulation remains policy-centric: the learned model is optimized to produce actions, and the reasoning that leads to those actions is largely internal to the policy. This can make the behavior difficult to diagnose or adapt when the scene changes, and recent robustness studies show that VLA policies remain sensitive to variations in object layout, viewpoint, robot initial state, lighting, background, physical conditions, and language prompts \cite{liu2025evavla,fei2025liberoplus,wang2025vlatest}. GTA-2 takes a complementary approach: instead of training an end-to-end action policy, it uses VLM agents to construct a structured zero-shot manipulation skill that can be grounded in the current scene, refined through feedback, and compiled into robot execution.

\vspace{-2mm}
\section{Method}

GTA-2 is a modular multi-step VLM framework that maps a natural-language instruction \(\ell\) and an RGB-D observation \(I\), together with a reusable controller library \(\mathcal{C}\), to an executable robot program \(P\). Rather than predicting low-level robot actions directly, GTA-2 constructs an explicit task-axis skill representation that connects language-level task reasoning to controller-level execution. The skill is generated without task-specific robot demonstrations, policy training, or fine-tuning.

The framework separates lifted/abstract skill synthesis from scene-specific grounding. A \textbf{Task Decomposer} converts the instruction into an ordered sequence of semantic subtasks. A \textbf{Skill Generator} then constructs a task-axis skill \(\tilde{\mathcal{S}}\) by selecting and composing controllers from \(\mathcal{C}\) and specifying the required keypoints, axes, controller priorities, and parameter placeholders. Scene-specific grounding comprises two complementary operations: a \textbf{Parameter Setter} assigns numerical values to these placeholders, while a \textbf{Vision Module} resolves the requested scene features to metric keypoints and axes. Together, these stages ground \(\tilde{\mathcal{S}}\) in the current scene, producing a grounded task-axis skill \(\mathcal{S}\). Fig.~\ref{fig:method_overview} summarizes these stages and the structured information passed between them.

A deterministic compiler converts the grounded skill into the executable program \(P\), whose task-axis controllers produce low-level position, orientation, force, and gripper commands through the robot execution interface. Because each stage exposes a separate intermediate output, users can refine the task decomposition, controller recipe, parameter assignments, or feature groundings without regenerating the entire behavior. The following sections first define the task-axis skill representation and then describe lifted skill synthesis, scene-specific grounding, robot execution, and targeted refinement.

\begin{table}[t]
\centering
\caption{Task-axis feature definitions.}
\label{tab:features}
\begin{tabularx}{\linewidth}{@{}p{0.35\linewidth}X@{}}
\toprule
\textbf{Feature} & \textbf{Description} \\
\midrule

\(k_{\mathrm{object\_part\_location}}\) &
A task-relevant keypoint on an object, identified by its object, part, and location. \\

\(k{_{eef}}{_{TCP}}\) &
Tool center point of the robot end-effector. \\

\(a{_{eef}}{_{X}},a{_{eef}}{_{Y}},a{_{eef}}{_{Z}}\) &
End-effector-frame axes for the orthogonal, pinch, and approach directions, respectively. \\

\(a{_{world}}{_{X}},a{_{world}}{_{Y}},a{_{world}}{_{Z}}\) &
Fixed world-frame axes for the front, right, and upward directions, respectively. \\

\(a_{surface\_normal}(k_1)\) &
Axis along normal pointing out of surface at keypoint specified by \(k_1\). \\

\(a_{edge\_tangent}(k_1)\) &
Axis along tangent direction of edge at keypoint specified by \(k_1\). \\

\(a_{along}(k_1, k_2)\) &
Axis pointing from keypoint \(k_1\) to keypoint \(k_2\). \\

\(a_{crossprod}(a_1, a_2)\) &
Unit-length axis computed by \(a_1  \times a_2\). \\

\bottomrule
\end{tabularx}
\end{table}

\subsection{Task-Axis Skill Representation}

GTA-2 builds on the task-axis skill representation introduced in Grounded Task-Axes~\cite{seker2025grounded}, in which manipulation behaviors are expressed as compositions of controller objectives grounded in task-relevant keypoints and axes. In GTA-2, this representation provides the intermediate structure that connects language-level task reasoning to controller-level robot execution.

For notation, we implicitly index the task-relevant objects in a scene as
\(
\mathcal{O}=\{O_i\}_{i=1}^{N},
\)
and associate each object \(O_i\) with a set of keypoint definitions
\(\mathcal{K}_i=\{k_{i,j}\}\) and axis definitions
\(\mathcal{A}_i=\{a_{i,m}\}\). A keypoint definition identifies a task-relevant location on or relative to an object, such as the center of a scribble, the tip of an eraser, or the rim of a container. An axis definition identifies a task-relevant direction, such as a surface normal, an object edge, or a direction between two keypoints. At the lifted level, these definitions specify which geometric features are required by the skill, but do not yet contain their metric 3D values. The sets include only the features relevant to the current skill and do not assume an exhaustive object inventory or a precomputed detection or segmentation of the scene. For notational simplicity, we omit the object index \(i\) in the remainder of the paper whenever the associated object is clear from context.

Table~\ref{tab:features} summarizes these scene-object features together with the robot-relative, world-relative, and derived feature definitions available to GTA-2. The same feature representation includes robot-relative and world-relative references. The robot end-effector is associated with keypoints and axes obtained from the tool frame, such as its tool center point and approach direction, while the world frame provides fixed reference directions such as the vertical axis. Axes may also be defined through deterministic geometric operations over other features, including a surface normal at a keypoint, the direction between two keypoints, or the cross product of two axes. These derived-axis definitions specify how the required direction should be computed once their input features receive metric values. 

A task-axis controller defines a parameterized control objective over the feature definitions introduced above. We represent a lifted controller as
\[
\tilde{c}
=
\left(
\tau,
\phi_{\mathrm{ctl}},
\phi_{\mathrm{ref}},
a_{\mathrm{axis}},
\tilde{\theta},
\rho
\right),
\]
where \(\tau\) denotes the controller type, \(\phi_{\mathrm{ctl}}\) is the robot or object feature being controlled, and \(\phi_{\mathrm{ref}}\) is the target feature relative to which the objective is defined. The optional task axis \(a_{\mathrm{axis}}\) specifies the direction along or about which the controller operates. The parameter placeholder \(\tilde{\theta}\) represents controller-specific quantities that have not yet been numerically assigned, while \(\rho\) specifies the controller's priority within the composition. The controller type \(\tau\) determines which of these fields are required and how they are interpreted.

Table~\ref{tab:controllers} summarizes the task-axis controller library used in our experiments. For example, a position-alignment controller uses control and target keypoints together with a task axis and positional offset, whereas a constant-force controller requires only a force direction and magnitude. Similarly, a gripper controller does not require control and target features, but instead specifies a gripper state and the time at which it should be applied. Thus, individual controller types may use only the subset of fields required by their respective objectives.

\begin{table}[t]
\centering
\caption{Task-axis controller library.}
\label{tab:controllers}
\begin{tabularx}{\linewidth}{@{}p{0.39\linewidth}X@{}}
\toprule
\textbf{Controller} & \textbf{Description} \\
\midrule

\textbf{PosAlign}\((k_1,k_2,a,o)\) &
Aligns \(k_1\) with \(k_2\) along \(a\) with offset \(o\). \\

\textbf{PosWaypoint}\((k_1,k_2,a,[o_n])\) &
Follows a sequence of positional offsets. \\

\textbf{AxisAlign}\((a_1,a_2,a_r,\phi)\) &
Aligns \(a_1\) with \(a_2\) by rotating about \(a_r\) with angular offset \(\phi\). \\

\textbf{AxisWaypoint}\((a_1,a_2,a_r,[\phi_n])\) &
Follows a sequence of angular offsets. \\

\textbf{ForceConstant}\((a,f)\) &
Maintains force value \(f\) along \(a\). \\

\textbf{GripperControl}\((s,t)\) &
Sets the gripper state \(s \in \{\text{``open''}, \text{``close''}\}\) at normalized subtask time \(t\). \\

\textbf{PosDMPSpiral}\((k,a,r)\) &
Reproduces a learned spiral of radius \(r\) at \(k\) in the plane orthogonal to \(a\). \\

\textbf{PosDMPLetter}\((k,a,l)\) &
Reproduces a specified letter trajectory in the plane orthogonal to \(a\), included as a distractor for task-relevant controller selection. \\

\bottomrule
\end{tabularx}
\end{table}

Multiple task-axis controllers can be composed to form a lifted skill step,
\(
\tilde{s}
=
[\tilde{c}_1,\ldots,\tilde{c}_M].
\)
Each controller contributes a distinct objective to the manipulation step, such as regulating position, orientation, force, motion, or gripper state. These objectives are evaluated in parallel and ordered according to their priorities \(\rho\). During execution, lower-priority control commands are projected into the null space of higher-priority controllers, preventing them from interfering with higher-priority objectives. A lifted skill step therefore describes both the individual objectives and the prioritized controller composition needed to realize a single phase of a manipulation task.

Consider the erasing step illustrated in Fig.~\ref{fig:teaser}. For compactness, we write
\(k_{\mathrm{scribble}}\) for
\(k_{\mathrm{redscribble\_middle\_center}}\)
and \(a_{\mathrm{normal}}\) for
\(a_{\mathrm{surface\_normal}}\).
Using the controller definitions in Table~\ref{tab:controllers}, the lifted erasing step may be composed as
\[
\tilde{s}_{\mathrm{erase}}
=
\left[
\begin{aligned}
&\operatorname{ForceConstant}
(a_{\mathrm{normal}}(k_{\mathrm{scribble}}),\tilde{f}),\\
&\operatorname{PosDMPSpiral}
(k_{\mathrm{scribble}},a_{\mathrm{normal}}(k_{\mathrm{scribble}}),\tilde{r})
\end{aligned}
\right].
\]
Here, \(\tilde{f}\) and \(\tilde{r}\) are parameter placeholders for the force magnitude and spiral radius, respectively. Relative controller priorities are implicit in the list order, with earlier controllers taking precedence over later ones. The composition remains lifted because its feature definitions and parameter placeholders have not yet been assigned scene-specific values.

A lifted controller becomes grounded when its feature definitions are resolved to metric values and its parameter placeholders are assigned numerical values. For the fields required by a given controller type, the grounded controller is written as
\[
c
=
\left(
\tau,
G(\phi_{\mathrm{ctl}}),
G(\phi_{\mathrm{ref}}),
G(a_{\mathrm{axis}}),
\theta,
\rho
\right),
\]
where \(G\) denotes feature grounding and \(\theta\) contains the assigned controller parameters.

Returning to the erasing example above, feature grounding provides the current scribble-center position and board-normal direction, while parameter assignment provides the desired force magnitude and spiral radius. Let:
\[
\begin{aligned}
G(k_{\mathrm{scribble}})
    &= \mathbf{p}_{\mathrm{scribble}}\in\mathbb{R}^{3}, &
G(a_{\mathrm{normal}})
    &= \mathbf{v}_{\mathrm{normal}}\in\mathbb{S}^{2},\\
\tilde{f} &\mapsto 3\,\mathrm{N}, &
\tilde{r} &\mapsto 5\,\mathrm{cm}.
\end{aligned}
\]
Applying these values yields the grounded erasing step
\[
s_{\mathrm{erase}}
=
\left[
\begin{aligned}
&\operatorname{ForceConstant}
(\mathbf{v}_{\mathrm{normal}},3\,\mathrm{N}),\\
&\operatorname{PosDMPSpiral}
(\mathbf{p}_{\mathrm{scribble}},
 \mathbf{v}_{\mathrm{normal}},
 5\,\mathrm{cm})
\end{aligned}
\right].
\]
The lifted and grounded skill steps preserve the same controller types, feature relationships, and priority ordering; only the metric feature values and numerical parameters are specific to the current scene. This separation allows the lifted controller composition to be reused across scene configurations while grounding adapts its execution to each observation.

\vspace{-2mm}
\subsection{Lifted Skill Synthesis}

Having defined the task-axis skill representation, we now describe how the first two stages of the GTA-2 pipeline, also shown in Fig.~\ref{fig:method_overview}, synthesize a lifted skill from the task instruction and scene context. The \textbf{Task Decomposer} and \textbf{Skill Generator} produce structured intermediate outputs rather than free-form robot code.

\paragraph*{\textbf{Task Decomposer}}
The Task Decomposer converts the natural-language instruction into an ordered sequence of semantic subtasks,
\[
\mathcal{Q}
=
A_{\mathrm{TD}}(\ell,I,\mathcal{C})
=
[q_1,\ldots,q_T],
\]
where each \(q_t\) describes a meaningful phase of the manipulation task without specifying the controllers used to execute it. In the bottle-placement example shown in Fig.~\ref{fig:method_overview}, the Task Decomposer separates the task into approaching the bottle, grasping it, lifting it, moving it over the box, placing it, and releasing the gripper. Similarly, for the erasing task in Fig.~\ref{fig:teaser}, it decomposes the behavior into alignment, contact, erasing, and retraction phases. These decompositions expose the temporal and semantic structure of each task and provide the scaffold from which the Skill Generator constructs the corresponding controller objectives.

\paragraph*{\textbf{Skill Generator}}
The Skill Generator maps each semantic subtask \(q_t\) to a lifted skill step,
\[
\tilde{s}_t
=
A_{\mathrm{SG}}(q_t,\ell,I,\mathcal{C})
=
[\tilde{c}_{t,1},\ldots,\tilde{c}_{t,M_t}].
\]
For each subtask, it selects the controller types from \(\mathcal{C}\), identifies the required keypoint and axis definitions, specifies the parameter placeholders, and arranges the resulting controllers in priority order. In the bottle-placement example in Fig.~\ref{fig:method_overview}, the \emph{Move over the box} subtask is mapped to position-alignment controllers that coordinate the bottle with the box interior and an axis-alignment controller that constrains the end-effector orientation. Similarly, for the erasing phase in Fig.~\ref{fig:teaser}, the Skill Generator produces the prioritized force-control and spiral-motion composition defined in Sec.~III-A. The resulting skill step remains lifted because its feature definitions have not yet been grounded and its parameter placeholders have not yet been assigned numerical values.

Applying the Skill Generator to each subtask yields the full lifted task-axis skill,
\[
\tilde{\mathcal{S}}
=
[\tilde{s}_1,\ldots,\tilde{s}_T].
\]
The ordering of the skill steps preserves the temporal structure established by the Task Decomposer, while the ordering of controllers within each \(\tilde{s}_t\) encodes their priorities. At this stage, \(\tilde{\mathcal{S}}\) specifies the controller types, feature definitions, parameter placeholders, and their composition, but does not yet contain metric feature groundings or numerical parameter values. The next stage resolves these remaining quantities for the observed scene.

\vspace{-2mm}
\subsection{Scene-Specific Skill Grounding}

The lifted skill synthesized above specifies what geometric relationships and controller objectives are required, but not the metric values needed for execution. GTA-2 resolves these quantities through two complementary operations: \emph{Feature grounding} maps each keypoint or axis definition to a 3D point or unit direction in the current execution context, whereas \emph{parameter assignment} replaces each parameter placeholder with a task- and scene-appropriate numerical value. We refer to the joint application of these operations to \(\tilde{\mathcal{S}}\) as \emph{scene-specific skill grounding}, which produces the grounded skill \(\mathcal{S}\) while preserving its controller composition and priority structure.

\paragraph*{\textbf{Parameter Setter}}
The Parameter Setter assigns numerical values to the parameter placeholders appearing in the lifted skill. Conditioned on the task instruction, scene observation, and controller context, it produces
\[
\Theta_{\ell,I}
=
A_{\mathrm{PS}}
(\ell,I,\tilde{\mathcal{S}},\mathcal{C})
=
\{\tilde{\theta}_j \mapsto \theta_j\}_{j=1}^{J},
\]
where each \(\tilde{\theta}_j\) is a placeholder in \(\tilde{\mathcal{S}}\) and \(\theta_j\) is its assigned numerical value. These assignments include controller-specific quantities such as positional and angular offsets, force magnitudes, trajectory scales, and execution timings. The Parameter Setter leaves the controller types, feature definitions, and priority ordering of the lifted skill unchanged.

\paragraph*{\textbf{Vision Module}}
The Vision Module resolves the feature definitions requested by \(\tilde{\mathcal{S}}\) to metric values for the current execution context. It produces the grounding map
\[
G_{I,\mathbf{x}}
=
A_{\mathrm{VG}}(I,\mathbf{x},\tilde{\mathcal{S}})
=
\{k_j\mapsto\mathbf{p}_j,\;a_m\mapsto\mathbf{v}_m\}.
\]
Here, \(k_j\) and \(a_m\) range over the keypoint and axis definitions requested by \(\tilde{\mathcal{S}}\), with \(\mathbf{p}_j\in\mathbb{R}^{3}\) and \(\mathbf{v}_m\in\mathbb{S}^{2}\), respectively. The state \(\mathbf{x}\) contains the robot state and calibrated frame information. Scene-object keypoints are localized in the RGB image by the VLM and recovered as metric 3D positions using the corresponding depth observation. Derived axes are evaluated from grounded features using point-cloud and geometric operations, while robot- and world-relative features are obtained from forward kinematics and calibrated fixed frames.

Together, the feature groundings and parameter assignments transform the full lifted skill into a grounded task-axis skill,
\[
\mathcal{S}
=
\bigl(
G_{I,\mathbf{x}},
\Theta_{\ell,I}
\bigr)
(\tilde{\mathcal{S}})
=
[s_1,\ldots,s_T],
\]
where each \(s_t\) is the grounded counterpart of the lifted skill step \(\tilde{s}_t\). This operation preserves the temporal ordering of the skill steps, the controller composition within each step, and their relative priorities, while replacing feature definitions and parameter placeholders with their corresponding metric and numerical values. In the erasing example introduced earlier, the Vision Module provides \(\mathbf{p}_{\mathrm{scribble}}\) and \(\mathbf{v}_{\mathrm{normal}}\), while the Parameter Setter supplies the force magnitude of \(3\,\mathrm{N}\) and spiral radius of \(5\,\mathrm{cm}\). Once all skill steps have been grounded, \(\mathcal{S}\) contains the complete controller specification required for compilation and execution.

\vspace{-2mm}
\subsection{Robot Script Compilation and Execution}

A deterministic compiler converts the grounded task-axis skill into an executable robot program,
\(
P
=
\Pi(\mathcal{S},\mathcal{C}).
\)
For each grounded controller in \(\mathcal{S}\), the compiler selects the corresponding implementation from the controller library \(\mathcal{C}\) and inserts its grounded features, numerical parameters, and priority information into the generated program. 

During execution, the program activates the grounded skill steps \(s_t\) in their specified temporal order. Within the active step, each controller evaluates its objective using the current robot state and the grounded feature values, and the resulting control commands are combined according to the prescribed priority ordering. The execution backend then transmits the resulting position, orientation, force, and gripper commands through the ROS-based robot interface. By confining robot-specific communication and command formatting to this backend, GTA-2 keeps lifted skill synthesis and scene-specific grounding independent of the low-level execution interface.


\vspace{-2mm}
\subsection{Human-Guided Targeted Refinement}

The structured intermediate outputs exposed by GTA-2 allow user feedback to be directed to the stage responsible for an observed error. As illustrated in Fig.~\ref{fig:method_overview}, decomposition errors can be routed to the Task Decomposer, controller or feature-definition errors to the Skill Generator, numerical parameter errors to the Parameter Setter, and geometric grounding errors to the Vision Module. For example, if the generated behavior moves too close to an object because of an inappropriate positional offset, the user can direct feedback specifically to the Parameter Setter. Given targeted feedback \(h_{\mathrm{PS}}\), the revised parameter assignments are
\[
\Theta'_{\ell,I}
=
A_{\mathrm{PS}}
(\ell,I,\tilde{\mathcal{S}},\mathcal{C},h_{\mathrm{PS}}).
\]
The updated grounded skill is then obtained as
\(
\mathcal{S}'
=
\bigl(
G_{I,\mathbf{x}},
\Theta'_{\ell,I}
\bigr)
(\tilde{\mathcal{S}}).
\)
Because the correction is confined to the Parameter Setter, the existing task decomposition, lifted controller composition, and feature groundings are reused; only the affected parameter assignments are revised.

More generally, targeted feedback is provided only to the agent responsible for the erroneous intermediate output. When a correction changes an upstream output, the downstream stages that depend on it are recomputed, while unaffected intermediate results are retained. GTA-2 can therefore preserve the correctly generated portions of a behavior while revising the component responsible for the failure, without policy retraining or indiscriminately regenerating the entire pipeline.

\begin{figure*}
    \centering
    \includegraphics[width=\linewidth]{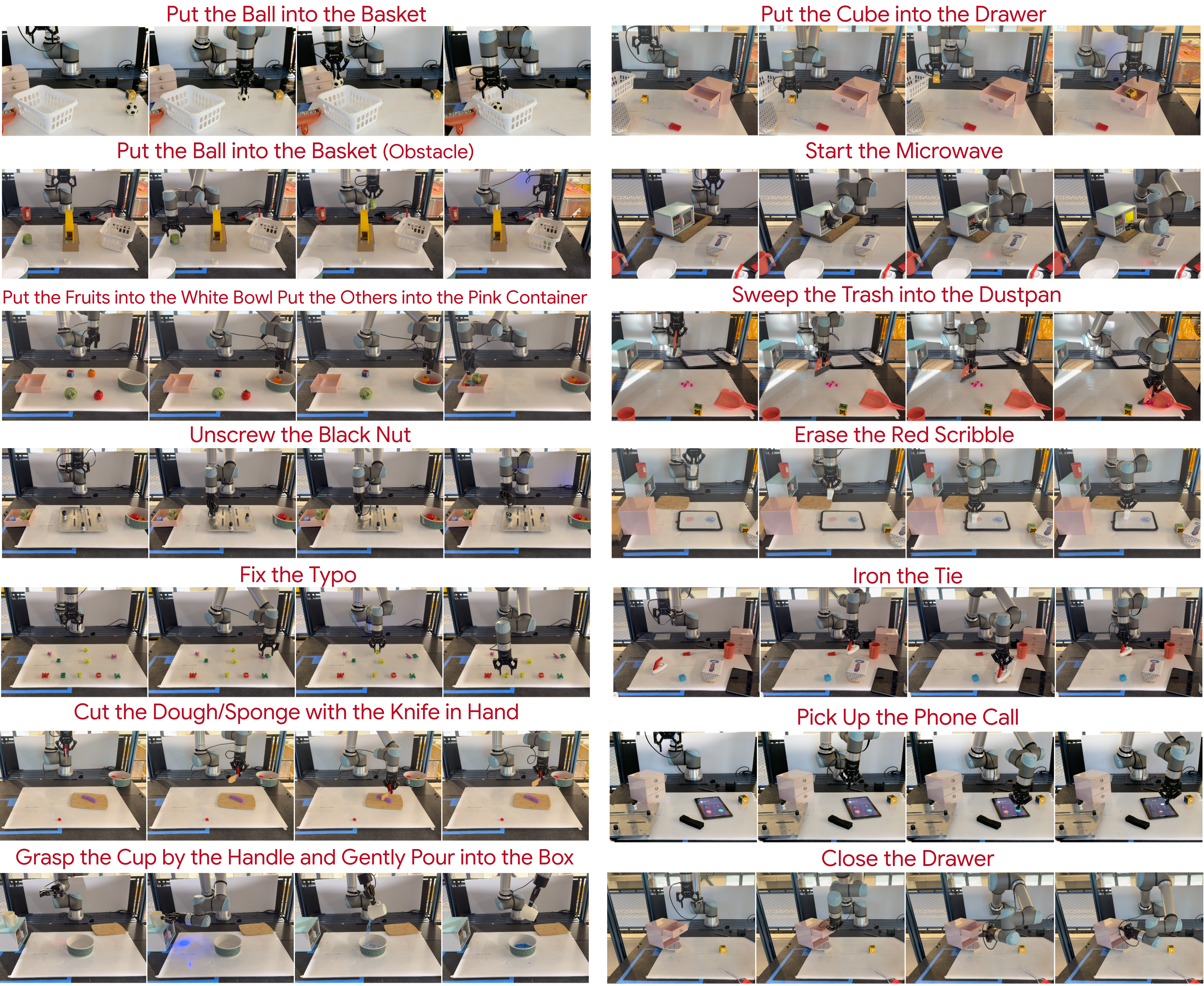}
    \caption{Representative real-robot executions across the 14 manipulation tasks used in our evaluation. Each example shows key aspects of the task diversity handled by GTA-2, including pick-and-place, articulated-object interaction, tool use, contact-rich manipulation, and constrained motion.}
    \vspace{-6mm}
    \label{fig:tasks}
\end{figure*}

\vspace{-2mm}
\section{Experimental Results}
We evaluate GTA-2 on 14 real-robot manipulation tasks spanning pick-and-place, articulated-object interaction, contact-rich manipulation, tool use, and constrained motion (Fig.~\ref{fig:tasks}). Our evaluation addresses two complementary questions. First, how reliably can GTA-2 generate executable skills without task-specific examples, and how effectively can agent-level feedback recover unsuccessful generations? Second, how well do the initial and refined skills generalize across scene configurations compared with VLA and Code-as-Policies baselines? We first describe the common experimental setup and comparison methods before presenting these two studies.

\subsection{Experimental Setup}

Table~\ref{tab:controllers} summarizes the controller library used in our experiments. Controllers may be composed concurrently within a subtask to express simultaneous position, orientation, force, and gripper objectives, and the same library is available across all evaluated tasks.

We use Gemini 3.1 Pro~\cite{gemini31pro} for all four VLM agents. The agents therefore share the same underlying model and differ only in their role-specific instructions, contextual inputs, and structured output representations. We use no task-specific demonstrations or model fine-tuning.

Physical experiments are conducted with a UR5e manipulator equipped with a Robotiq 2F-85 parallel-jaw gripper and a fixed ZED 2i stereo camera calibrated to the robot base frame. RGB observations provide visual context for the VLM agents, while the corresponding point cloud supplies metric 3D keypoints and geometrically derived task axes, including surface normals and directions between grounded keypoints.

\subsection{Baseline Methods}
We compare GTA-2 against an end-to-end vision-language-action policy and two Code-as-Policies (CaP) baselines. All methods receive the same task instructions and scene observations.

\paragraph*{\textbf{VLA baseline ($\pi_{0.5}$)}}
We use the $\pi_{0.5}$ DROID-finetuned version as a representative generalist VLA policy for open-world manipulation \cite{intelligence2025pi_,Khazatsky-RSS-24}. Like GTA-2, it uses language and vision to generate high-level subtasks, but it ultimately follows an end-to-end formulation that maps observations and instructions to robot actions. We implemented this baseline on the Franka Emika robot for improved performance.

\paragraph*{\textbf{Task-axis controller baseline (CaP-TAC)}}
CaP-TAC is a Code-as-Policies-style baseline in which a VLM directly generates executable robot policy code from the task instruction and scene observation \cite{liang2023code}. To isolate the contribution of GTA-2's structured abstraction-to-grounding pipeline, CaP-TAC is given access to the same sub-atomic task-axis controllers. The generated program may compose multiple controllers concurrently into multiple subtasks. However, CaP-TAC must select perceptual queries, determine the required task geometry, and compose the controller sequence directly within a single generated robot script, without GTA-2's intermediate task abstractions or structured grounding procedure.

\paragraph*{\textbf{Primitive-action baseline (CaP-Primitive)}}
CaP-Primitive follows the same code-generation and perception protocol as CaP-TAC but replaces task-axis controllers with three conventional robot primitives: absolute Cartesian TCP motion, gripper actuation, and Cartesian motion under a specified contact force. This baseline tests whether direct policy-code generation is sufficient when provided with standard robot primitives, while comparison with CaP-TAC isolates the effect of exposing task-axis controllers.

Both CaP baselines use the same policy-generation model and the vision module (Gemini-3.1-Pro) as GTA-2. Both baselines are supported by a few-shot setting using three fixed demonstrations covering pick-and-place, force-controlled surface contact, and point traversal. The demonstrations are matched in task content across the two APIs and do not include the benchmark tasks themselves

\begin{figure*}
    \centering
    \includegraphics[width=\linewidth]{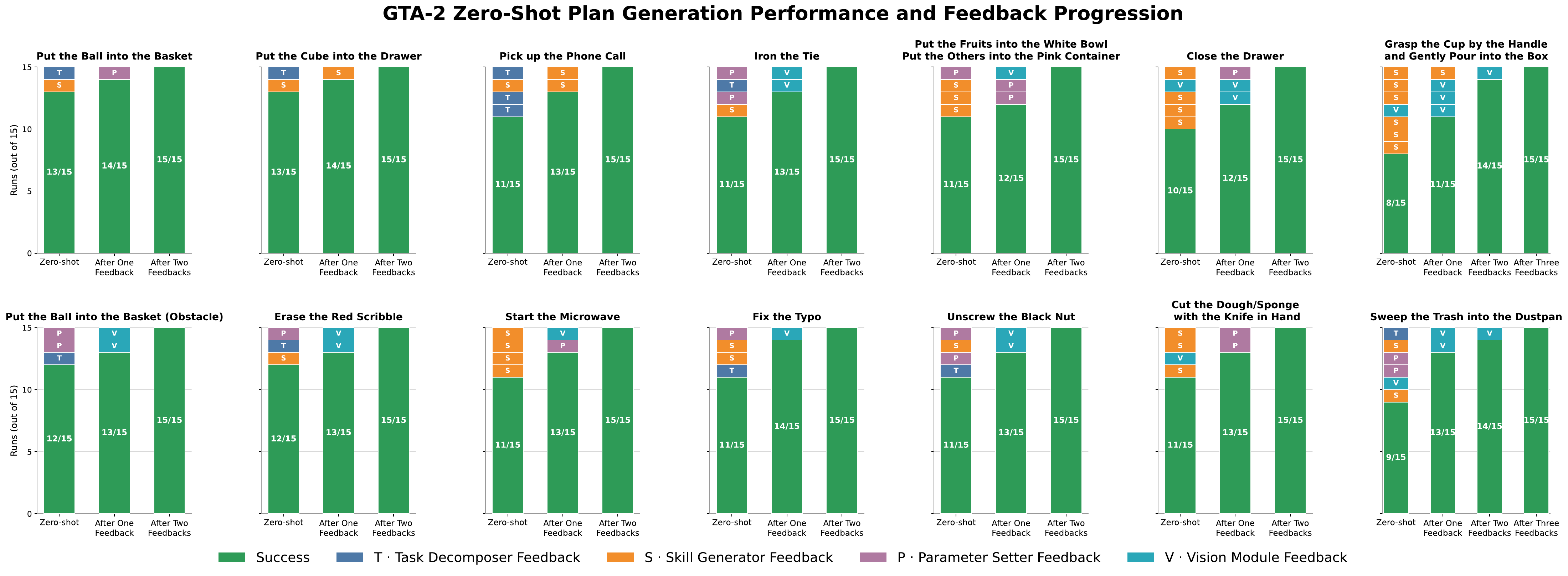}
    \caption{\textbf{Zero-shot plan-generation performance and feedback-based recovery of GTA-2 across 14 manipulation tasks.} Each task is evaluated over 15 independent runs. Green segments indicate successful executions, while the remaining segments denote runs requiring feedback at that stage and are colored by the module receiving feedback.}
    \vspace{-4mm}
    \label{fig:feedback_progression}
\end{figure*}
\vspace{-2mm}

\subsection{Zero-Shot Generation and Targeted Refinement}

We first evaluate two central properties of GTA-2: its ability to generate executable manipulation skills without task-specific examples and its ability to refine unsuccessful generations through localized feedback. For each of the 14 tasks, we perform 15 skill-generation runs and validate the resulting skills through physical execution. Each run begins without task-specific feedback. When an execution reveals an error, feedback is directed only to the responsible module, while intermediate representations that remain valid are preserved. The experiment therefore evaluates structured refinement rather than repeated unconstrained regeneration of the complete skill.

As shown in Fig.~\ref{fig:feedback_progression}, GTA-2 succeeds without feedback in 154 of the 210 independent runs ($73.3\%$), including tasks involving sustained contact and tool use. Although pouring and sweeping require more refinement than direct transport behaviors, their successful zero-shot runs demonstrate that the generated task-axis representation can express these behaviors without task demonstrations or manually specified skill structures. Human feedback increases the cumulative number of successful runs to 181/210 ($86.2\%$) after one round and 208/210 ($99.0\%$) after two rounds, with all 210 runs succeeding within three. Task complexity therefore increases the amount of refinement required, but does not require new primitives, training, or reconstruction of the complete skill. A small number of high-level corrections is sufficient to turn independently generated failures into successful executions.

\begin{figure*}
    \centering
    \includegraphics[width=\linewidth]{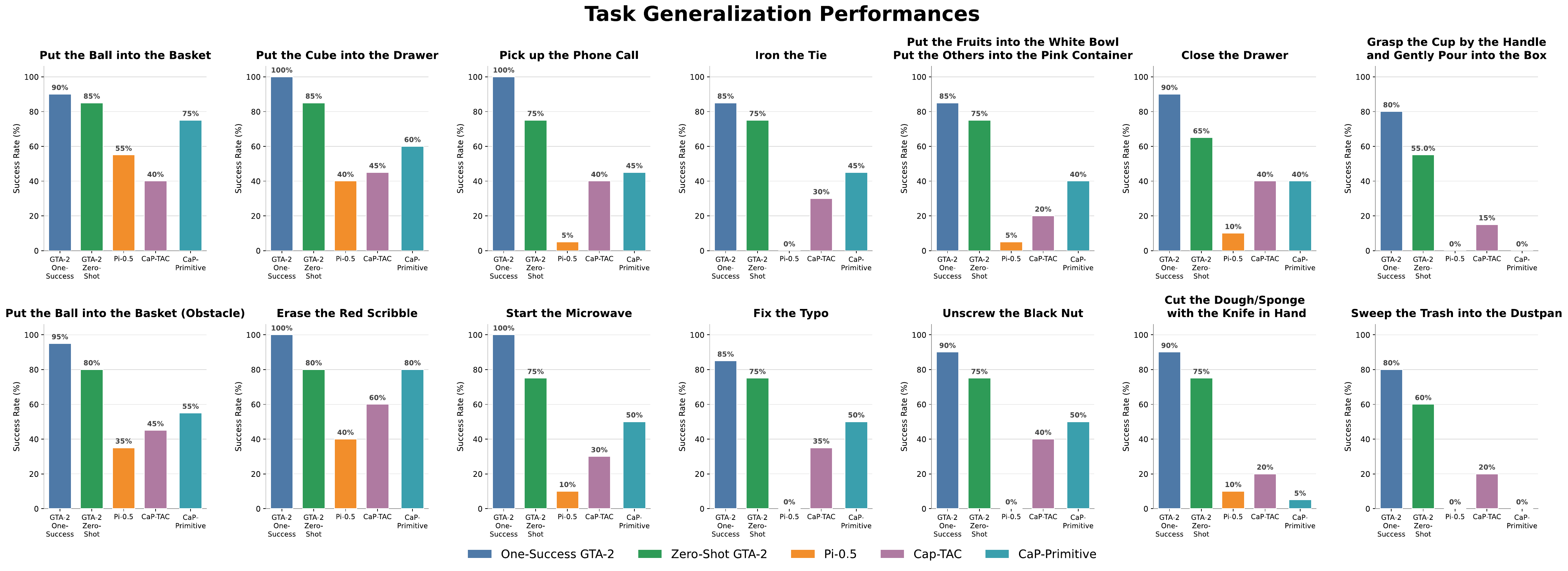}
    \caption{\textbf{Task-generalization performance of GTA-2 and the evaluated baselines across 14 manipulation tasks.} Bars report success rates for GTA-2 under the One-Success and Zero-Shot settings, together with Pi-0.5, CaP-TAC, and CaP-Primitive. All results are presented on a common 0–100\% scale and evaluated over 20 runs. Values above the bars show the corresponding success rates.}
    \vspace{-6mm}    
    \label{fig:task_generalization}
\end{figure*}

The feedback distribution further shows that GTA-2's agent boundaries correspond to meaningful and independently correctable failure modes. Initial feedback is directed primarily to the Skill Generator, indicating that the task is generally interpreted correctly and that the main remaining challenge is translating this intent into an appropriate task-axis controller composition. Once these structural decisions are corrected, the remaining feedback shifts predominantly to the Vision Module, while no further abstract planning corrections are required. This ordered transition mirrors GTA-2's abstraction-to-grounding pipeline: it establishes manipulation structure first, after which residual scene-specific grounding errors can be isolated and corrected. The specialized agents therefore provide not only interpretable intermediate outputs but also explicit refinement interfaces, allowing each correction to address the current error source without discarding decisions already resolved by other stages.

\subsection{Task Generalization and Baseline Comparison}

We next evaluate task generalization across 14 manipulation tasks and varying scene configurations, using 20 trials per task and method. The GTA-2 results correspond to two stages of the preceding refinement experiment. \textit{GTA-2 Zero-Shot} uses the initial skill before feedback, shown by the first bar for each task in Fig.~\ref{fig:feedback_progression}. \textit{GTA-2 One-Success} uses the refined skill after feedback has produced a successful execution, shown by the final bar. In both conditions, the resulting skill structure is retained and regrounded for each evaluation scene. We compare them with $\pi_{0.5}$, \textit{CaP-TAC}, and \textit{CaP-Primitive} under the common protocol described above.

As shown in Fig.~\ref{fig:task_generalization}, GTA-2 Zero-Shot achieves an average success rate of $73.9\%$, exceeding the strongest baseline, CaP-Primitive at $42.5\%$, by $31.4$ percentage points. GTA-2 One-Success further raises the average success rate to $90.7\%$. The improvement remains effective over subsequent scene configurations, indicating that refinement corrects reusable components of the skill rather than only the execution in which feedback was provided.

CaP-Primitive illustrates both the appeal and the limitation of monolithic primitives. By packaging common manipulation behaviors into a small number of function calls, they simplify planning and can perform well when a task matches the expected geometric pattern, particularly for top-down manipulation. However, action sequencing, parameter selection, and visual grounding must still be determined jointly within a single generated program. This coupling makes the resulting policy sensitive to small errors in the planning aspect of program generation.  GTA-2 instead assigns these decisions to specialized agents and communicates their outputs through explicit intermediate representations. Its advantage therefore comes not from simplifying the available actions, but from reducing the coupling between reasoning stages.

CaP-TAC receives the same task-axis controller interface as GTA-2 but must select, parameterize, ground, and compose the controllers within a single program. Its lower overall performance than CaP-Primitive is therefore expected: task-axis controllers provide a richer action space but also require more structured reasoning. Nevertheless, CaP-TAC outperforms CaP-Primitive on pouring, cutting, and sweeping, where execution depends on contact constraints, object-relative motion, and tool geometry. These results identify two complementary contributions: task-axis representations provide an effective abstraction for contact-rich manipulation and tool use, while GTA-2's specialized multi-agent pipeline makes this abstraction reliable by separately resolving task decomposition, controller composition, parameterization, and visual grounding.

\vspace{-2mm}
\section{Limitations and Future Work}
Although GTA-2 generates manipulation skills without task-specific demonstrations, skill synthesis remains constrained by the expressivity of the task-axis representation and the available controller library. GTA-2 can generate task-specific keypoint and axis definitions, but objectives that cannot be expressed through the available interface require extending its feature or controller libraries. Moreover, the sparse geometric representation assumes that the required scene-object features can be observed or inferred from a single RGB-D observation. These constraints primarily reflect the current instantiation of GTA-2: its modular pipeline could accommodate additional feature definitions, controller types, and sensing modalities without changing its overall structure.

Scene-object features are currently grounded before execution and are not updated online. While the task-axis controllers operate in closed loop with respect to the robot state and force measurements, the feature groundings and subtask structure are not continuously revised during execution. Object motion, slippage, or other contact-induced scene changes that invalidate the initial groundings therefore require a new grounding or refinement cycle before re-execution. The observed shift of feedback toward the Vision Module in later refinement rounds further indicates that perceptual grounding becomes an important residual source of failure once the manipulation structure has been established.

Our physical evaluation uses a single tabletop robot, gripper, and sensing configuration. Although the task-axis representation separates generated skills from the robot-specific execution backend, experiments across additional embodiments and sensing configurations are needed to establish its broader generality. Future work will also investigate an orchestrator VLM that monitors execution observations, diagnoses failures, identifies the stage requiring revision, and directs targeted feedback to the corresponding specialized agent. Such an orchestrator could reduce reliance on human feedback and enable an autonomous process for iterative execution monitoring and skill refinement.

\vspace{-4mm}
\section{Conclusion}
We presented GTA-2, a modular multi-VLM framework that bridges language-level task intent and robot execution through an explicit object-centric task-axis skill representation. By separating task decomposition, controller composition, parameter setting, and visual grounding, GTA-2 constructs executable manipulation behaviors from reusable controllers without task-specific demonstrations, policy training, or fine-tuning. This structure also makes the generated behavior interpretable and enables human feedback to be directed to the responsible stage while preserving decisions that are already correct.

Across 14 diverse real-robot manipulation tasks, GTA-2 demonstrated stronger and more consistent zero-shot performance than the evaluated VLA and Code-as-Policies baselines. Targeted refinement efficiently recovered unsuccessful generations and produced skills that remained effective across subsequent scene configurations. The results further show that task-axis controllers are most effective when paired with specialized reasoning stages and explicit intermediate representations. Together, these findings establish GTA-2 as a practical step toward open-world manipulation systems that can construct, inspect, and adapt new behaviors rather than relying on fixed skill primitives or opaque end-to-end policies.

\bibliographystyle{IEEEtran}
\bibliography{references}


\end{document}